\documentclass[conference]{IEEEtran}

\usepackage{multirow}
\usepackage{graphicx}
\usepackage{hyperref}

\usepackage{caption}
\IEEEoverridecommandlockouts

\usepackage{tikz}
\usepackage{textcomp}
\usepackage{hyperref}
\usepackage{lipsum}

\ifCLASSINFOpdf
\else
\fi
\begin{document}
%
\title{Fully Convolutional Networks for Diabetic Foot Ulcer Segmentation}


\author{\IEEEauthorblockN{Manu Goyal\IEEEauthorrefmark{1},
Neil D. Reeves\IEEEauthorrefmark{2},
Satyan Rajbhandari\IEEEauthorrefmark{3}, 
Jennifer Spragg\IEEEauthorrefmark{4} and 
Moi Hoon Yap\IEEEauthorrefmark{1}}
\IEEEauthorblockA{\IEEEauthorrefmark{1}School of Computing, Mathematics and Digital Technology\\
	Manchester Metropolitan University, Manchester, UK, M1 5GD\\
	Email: m.yap@mmu.ac.uk§}
\IEEEauthorblockA{\IEEEauthorrefmark{2}School of Healthcare Sciences\\
	Manchester Metropolitan University, Manchester, UK, M1 5GD}
\IEEEauthorblockA{\IEEEauthorrefmark{3}Lancashire Teaching Hospital, Preston, UK, PR2 9HT}
\IEEEauthorblockA{\IEEEauthorrefmark{4}Lancashire Care NHS Foundation Trust, Preston, UK, PR5 6AW}}


%


\IEEEspecialpapernotice{\textcopyright 2017 IEEE. Personal use of this material is permitted. Permission from IEEE must be obtained for all other uses, in any current or future media, including reprinting/republishing this material for advertising or promotional purposes, creating new collective works, for resale or redistribution to servers or lists, or reuse of any copyrighted component of this work in other works.}

\maketitle
\def\maketitle{\par%
	\begingroup%
	\normalfont%
	\def\thefootnote{}
	\def\footnotemark{}
	\let\@makefnmark\relax
	\footnotesize
	\footnotesep 0.7\baselineskip
	\@IEEEcompsoconly{\long\def\@makefntext##1{\parindent 1em\noindent\hbox{\@makefnmark}##1}}%
	\normalsize%
	\ifCLASSOPTIONpeerreview
	\newpage\global\@topnum\z@ \@maketitle\@IEEEstatictitlevskip\@IEEEaftertitletext%
	\thispagestyle{IEEEpeerreviewcoverpagestyle}\@thanks%
	\else
	\if@twocolumn%
	\ifCLASSOPTIONtechnote%
	\newpage\global\@topnum\z@ \@maketitle\@IEEEstatictitlevskip\@IEEEaftertitletext%
	\else
	\twocolumn[\@maketitle\@IEEEdynamictitlevspace\@IEEEaftertitletext]%
	\fi
	\else
	\newpage\global\@topnum\z@ \@maketitle\@IEEEstatictitlevskip\@IEEEaftertitletext%
	\fi
	\thispagestyle{IEEEtitlepagestyle}\@thanks%
	\fi
	\if@IEEEusingpubid
	\enlargethispage{-\@IEEEpubidpullup}%
	\fi 
	\endgroup
	\setcounter{footnote}{0}\let\maketitle\relax\let\@maketitle\relax
	\gdef\@thanks{}%
	\let\thanks\relax
}
\begin{abstract}
Diabetic Foot Ulcer (DFU) is a major complication of Diabetes, which if not managed properly can lead to amputation.
DFU can appear anywhere on the foot and can vary in size, colour, and contrast depending on various pathologies.
Current clinical approaches to DFU treatment rely on patients and clinician vigilance, which has significant limitations such as the high cost involved in the diagnosis, treatment and lengthy care of the DFU. We introduce a dataset of 705 foot images. We provide the ground truth of ulcer region and the surrounding skin that is an important indicator for clinicians to assess the progress of ulcer. Then, we propose a two-tier transfer learning from bigger datasets to train the Fully Convolutional Networks (FCNs) to automatically segment the ulcer and surrounding skin. Using 5-fold cross-validation, the proposed two-tier transfer learning FCN Models achieve a Dice Similarity Coefficient of 0.794 ($\pm$0.104) for ulcer region, 0.851 ($\pm$0.148) for surrounding skin region, and 0.899 ($\pm$0.072) for the combination of both regions. This demonstrates the potential of FCNs in DFU segmentation, which can be further improved with a larger dataset.
\end{abstract}


%
\IEEEpeerreviewmaketitle

\section{Introduction}
Diabetes Mellitus (DM) commonly known as Diabetes, is a serious and chronic metabolic disease that is characterized by elevated blood glucose due to insufficient insulin produced by the pancreas (Type 1) and human body's inability to use insulin effectively (Type 2) \cite{jeffcoate2003diabetic}. It can further causes major life-threatening complications like potential blindness, cardiovascular, peripheral vascular and cerebrovascular diseases, kidney failure and Diabetic Foot Ulcers (DFU) which can lead to lower limb amputation \cite{wild2004global}. There is a meteoric rise in diabetes from 108 million patients to 422 million worldwide where the low/middle income countries are disproportionately affected. In 2012, over 1.5 million deaths were caused by diabetes only and 43\% of these deaths are under the age of 70 years. It is estimated that by the end of 2035, around 600 million people will be suffering from DM \cite{bakker20162015}. Every year, more than 1 million patients suffering from diabetes lose part of their the leg due to the failure to recognize and treat DFU appropriately \cite{armstrong1998validation}. 

In current practice, medical experts (DFU specialist and podiatrist) primarily examine and assess the DFU patients on visual inspection with manual measurements tools to determine the severity of DFU. They also use the high-resolution images to evaluate the state of DFU, which can further comprise of various important tasks in early diagnosis, keeping track of development and number of actions taken to treatment and management of DFU for each particular case: 1) the medical history of patient is evaluated; 2) a wound or DFU specialist examines the DFU thoroughly; 3) additional tests like CT scans, MRI, X-Ray may be useful to help develop a treatment plan \cite{edmonds2006diabetic}. Usually, the DFU have irregular structures and uncertain outer boundaries. The appearance of DFU and its surrounding skin varies depending upon the various stages i.e. redness, callus formation, blisters, significant tissues types like granulation, slough, bleeding, scaly skin \cite{lipsky2004diagnosis}. 

The skin surrounding around DFU is very important as its condition determines if the DFU is healing and is also a vulnerable area for extension \cite{steed1996effect,rajbhandari1999digital}. There are many factors that increase the risk of vulnerable skin such as ischemia, inflammation, abnormal pressure, maceration from exudates etc. Similarly, healthy skin around the DFU indicates good healing process. Surrounding skin is examined by inspection of colour, discharge and texture, and palpation for warmth, swelling and tenderness. On visual inspection, redness is suggestive of inflammation, which is usually due to wound infection. Black discoloration is suggestive of ischemia. White and soggy appearance is due to maceration and white and dry is usually due to increased pressure. It is important to recognize that skin appearances look different in different shades of skin. Lesions that appear red or brown in white skin, may appear black or purple in black or brown skin. Mild degrees of redness may be masked completely in dark skin.

\begin{figure*}
	\centering
	\includegraphics[scale=0.6]{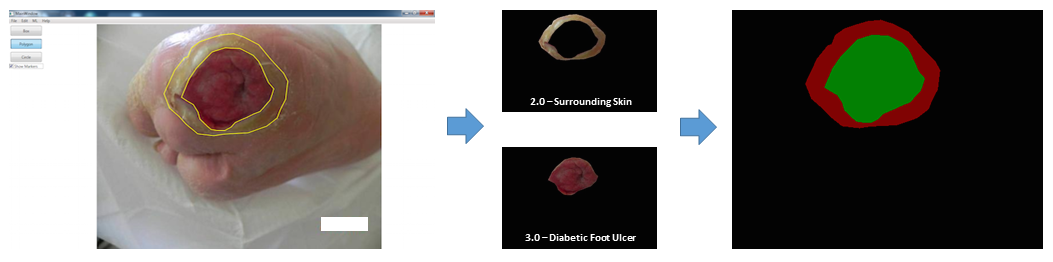}
	\caption{An example of delineating the different regions of the pathology from the whole foot image and conversion to Pascal VOC format}
	\label{fig:Annotator}
\end{figure*}

There are limited computer methods developed for the assessment of diabetic foot pathologies \cite{liu2015automatic,wang2016area}. Wang et. al. \cite{wang2016area} have used an image capture box to capture image data and determined the area of DFU using cascaded two staged SVM-based classification. Similarly,  computer methods based on manually engineered features or image processing approaches were implemented for segmentation of DFU and wound. The segmentation task was performed by extracting texture descriptors and colour descriptors on small patches of wound images, followed by machine learning algorithms to classify them into normal and abnormal skin patches \cite{kolesnik2005multi,kolesnik2006robust,papazoglou2010image,veredas2010binary}. As in many computer vision systems, the hand-crafted features are affected by skin shades, illumination, and image resolution. Also, these techniques struggled to segment the irregular contour of the DFU or wounds. Additionally, due to the complication of surrounding skin, it is almost impossible to include the broad description of surrounding skin. On the other hand, the unsupervised approaches rely upon image processing techniques, edge detection, morphological operations and clustering algorithms using different colour space to segment the wounds from images \cite{yadav2013segmentation,castro2006analysis,chung2000segmenting}. The majority of these methods involve manually tuning of the parameters according to the different input images which are very impractical in clinical perspective. In addition to the limitations of the segmentation algorithms, the state-of-the-art methods were validated on relatively small datasets, ranging from 10 to 172 images.

In this work, we propose automated segmentation of DFU and its surrounding skin by using fully connected networks. The contributions of this paper include
\begin{enumerate}
	\item To overcome the deficiency of DFU dataset in the state of the art, we present the largest DFU dataset alongside with the annotated ground truth.
	\item This is the first attempt in computer vision methods to segment the significant surrounding skin separately from the DFU.
	\item We propose a two-tier transfer learning method by training the fully convolutional networks (FCNs) on larger datasets of images and use it as pre-trained model for the segmentation of DFU and its surrounding skin. The performance is compared to other deep learning framework and the state-of-the-art DFU/wound segmentation algorithms on our dataset.
\end{enumerate}
\section{Method}
This section describes the preparation of the dataset, this includes expert labelling of the DFU and surrounding skin on foot images. The description of segmentation using conventional methods and deep learning methods are detailed. Finally, the performance metrics used for validation are reported.

\begin{figure*}
	\centering
	\includegraphics[width=15cm,height=4.2cm]{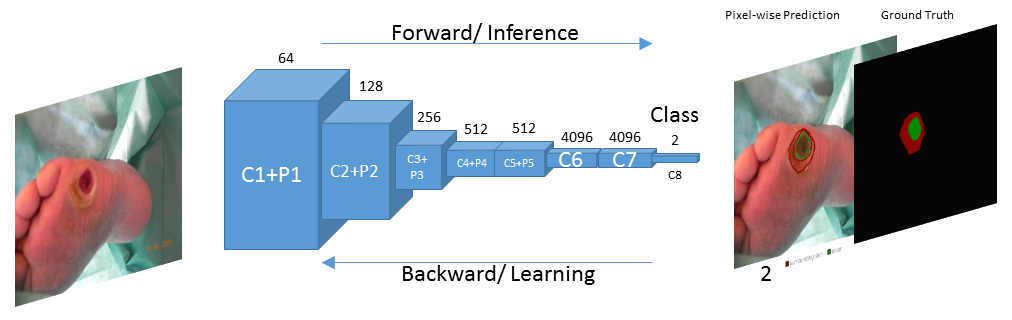}
	\caption{Overview of fully convolutional network architecture that learn features with forward and backward learning for pixel-wise prediction to perform segmentation where \textit{C1-C8} are convolutional layers and \textit{P1-P5} are max-pooling layers}
	\label{fig:Modelpredict}
\end{figure*}

\subsection{DFU Dataset}
A DFU dataset was collected over a five period at the Lancashire Teaching Hospitals and all the patients gave written informed consent. A subset of the images was used for this study, which include 600 DFU images and 105  healthy foot images. We received the NHS Research Ethics Committee approval with REC reference number 15/NW/0539 to use these images for our research. These DFU images were captured with Nikon D3300. Whenever possible, the images were acquired with close-ups of the full foot with the distance of around 30-40 cm with the parallel orientation to the plane of an DFU. The use of flash as the primary light source was avoided and instead, adequate room lights are used to get the consistent colours in images. To ensure the close range focus and avoiding the blurriness in images from the close distance, a Nikon AF-S DX Micro NIKKOR 40mm f/2.8G lens was used.

The ground truth annotation of our dataset was performed by a podiatrist specializing in the diabetic foot and validated by a consultant specializing in diabetes. We created ground truth for each image with DFU by using Hewitt et al. \cite{hewitt2016manual} annotator. For each DFU image (as illustrated in Fig.  \ref{fig:Annotator}), the expert delineated the region of interest (ROI) as the combination of DFU and its surrounding skin. Then in each ROI, the two classes were labelled separately and exported to an XML file. These ground truths were further converted into the label image of single channel 8-bit paletted image (commonly known as Pascal VOC format for semantic segmentation) as shown in Fig. \ref{fig:Annotator}. In this format, index 0 maps to black pixels represent the background, index 1 (red) represents the surrounding skin and index 2 (green) as DFU. From 600 DFU images in our dataset, we produce 600 ROIs of DFU and 600 ROIs for surrounding skin around the DFU.

\subsection{Fully Convolutional Networks for DFU segmentation}
Deep learning models proved to be powerful algorithms to retrieve hierarchies of features to achieve various tasks of computer vision. These convolutional neural networks, especially classification networks have been used to classify various classes of objects by assigning discrete probability distribution for each class. But, these networks have limitations as they are not able to classify multiple classes in a single image and figure out the position of the objects in images. FCNs instead produce segmentation by addressing these limitations by pixel-wise prediction rather than single probability distribution in the classification task for each image. Therefore, each pixel of a image is predicted for which class it belongs. The working of FCN architecture to produce pixel-wise prediction with the help of supervised pre-training using the ground truth is illustrated in Fig. \ref{fig:Modelpredict}. Hence, these models have the ability to predict multiple objects of various classes and position of each object in images. 

\begin{figure*}
	\centering
	\includegraphics[scale=0.5]{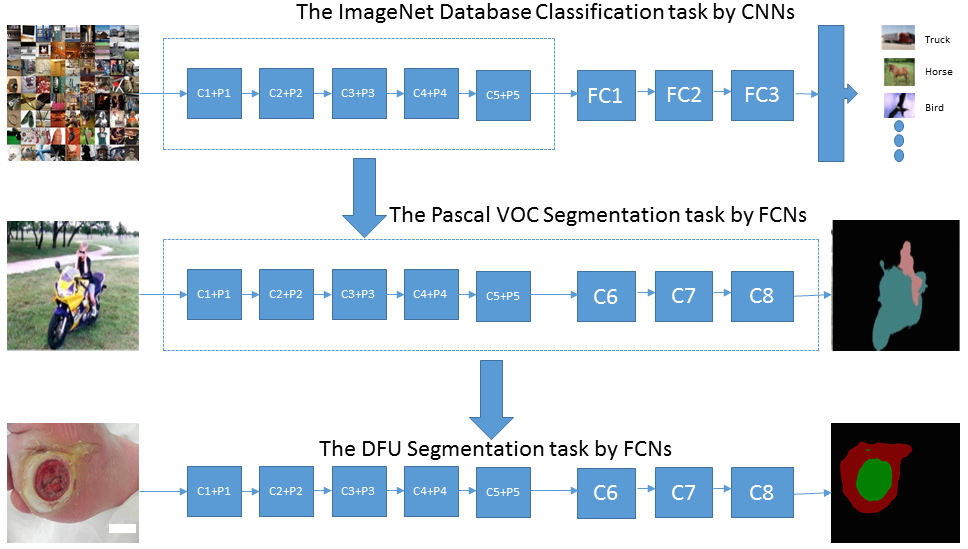}
	\caption{The two-tier transfer learning from big datasets to produce more effective segmentation}
	\label{fig:Transferlearning}
\end{figure*}
\subsection{Transfer Learning}
We used two-tier transfer learning for FCNs to perform more effective segmentation on DFU dataset. In first tier transfer learning, relevant CNN models that are used to make FCNs are trained on the ImageNet dataset with millions images \cite{ILSVRC15} for fine-tuning the weights associated with initial convolutional layers. In second tier transfer learning, we trained the FCN models on the Pascal VOC segmentation dataset \cite{Everingham10}. These pre-trained models are used for training the FCN models on DFU dataset for better convergence of weights associated with all layers of network rather than random initialization of weights. The two-tier transfer learning is illustrated in Fig. \ref{fig:Transferlearning}.    


\subsubsection{FCN-AlexNet}
The FCN-AlexNet is a fully convolutional network version of original classification model AlexNet by few adjustment of layers of networks for segmentation \cite{Long2015}. This network was originally used for classification of 1000 different objects of classes on the ImageNet dataset. It emerged as winner of ImageNet ILSVRC-2012 competition in classification category by achieving 99\% confidence \cite{krizhevsky2012imagenet}. There are few customizations made in the classification network model in order to convert it into FCN to carry out dense prediction. In FCN-AlexNet, earlier CNN layers are kept the same to extract the features and fully connected layers which throw away the positional coordinates are convolutionalized with the equivalent convolutional layers by adjusting the size of filters according to the size of the input to these layers \cite{Long2015}.  After the extraction of coarser and high-level features from input images, to produce the pixel-wise prediction for every pixel of the input, the deconvolutional layers work exactly opposite to the convolutional layers and stride used in this layer is equal to the scaling factor used in the convolutional layers. 

The input was 500$\times$500 foot images and ground truth images (Pascal VOC format). In the end, the network prediction on test images was very close to the ground truth. We used the Caffe \cite{jia2014caffe} framework to implement FCN-AlexNet. We have used these network parameters to train a model on the dataset i.e. 60 epochs, a learning method as stochastic gradient descent as rate of 0.0001 with a step-down policy and step size of 33\%, and gamma is 0.1. The learning parameter is decreased by the factor of 100 due to the introduction of new convolutional layers instead of fully connected layers which result in improved performance of FCN-AlexNet and other FCNs. 
\subsubsection{FCN-32s, FCN-16s, FCN-8s}
FCN-32s, FCN-16s, and FCN-8s are three models inspired by the VGG-16 based net which is a 16 layer CNN architecture that participated in the ImageNet Challenge 2014 and secured the first position in localization and second place in classification competition \cite{simonyan2014very,Long2015}. These models are customized with the different upsampling layers that magnify the output used in the original CNN model VGG-16. FCN-32s is same as of FCN-VGG16 in which fully connected layers are convolutionized and end to end deconvolution is performed with 32-pixel stride. The FCN-16s and FCN-8s additionally work on low-level features in order to produce more accurate segmentation. In FCN-16s, the final output is sum of upsampling of two layers i.e. upsampling of pool4 and 2$\times$upsampling of convolutional layer 7 whereas in FCN-8s, it is the sum of upsampling of pool3, 2$\times$upsampling of pool4 and 4$\times$upsampling of convolutional layer 7. Both models perform prediction on much more finer grained analysis i.e. 16$\times$16 pixel blocks for FCN-16s and 8$\times$8 pixel blocks for FCN-8s. The suitable pre-trained models for each model are also used in the training. The same input images are used to train the model with same parameters as of FCN-AlexNet i.e. 60 epochs, a learning rate of 0.0001, and gamma of 0.1.

\section{Experiment and Result}
As mentioned previously, we used the deep learning models for the segmentation task. The experiments were carried out on the DFU dataset of 600 DFU foot images that was splitted into the 70\% training, 10\% validation and 20\% testing. We adopted 5-fold cross-validation. For training and validation using the deep learning architecture, we used 420 images and 60 images respectively from the 600 original DFU images. Finally, we tested our model predictions on 120 remaining images. Further, we tested the performance of the models on 105 healthy test images. 

The performance evaluation of the FCN frameworks on the testing
set is achieved with 3 different DFU regions due to the practical medical applications. The DFU regions are explained below:
\begin{enumerate}
	\item The complete area determination (including Ulcer and Surrounding Skin).
	\item The DFU region
	\item The surrounding skin (SS) region
\end{enumerate}
\begin{figure*}
	\centering
	\includegraphics[scale=0.6]{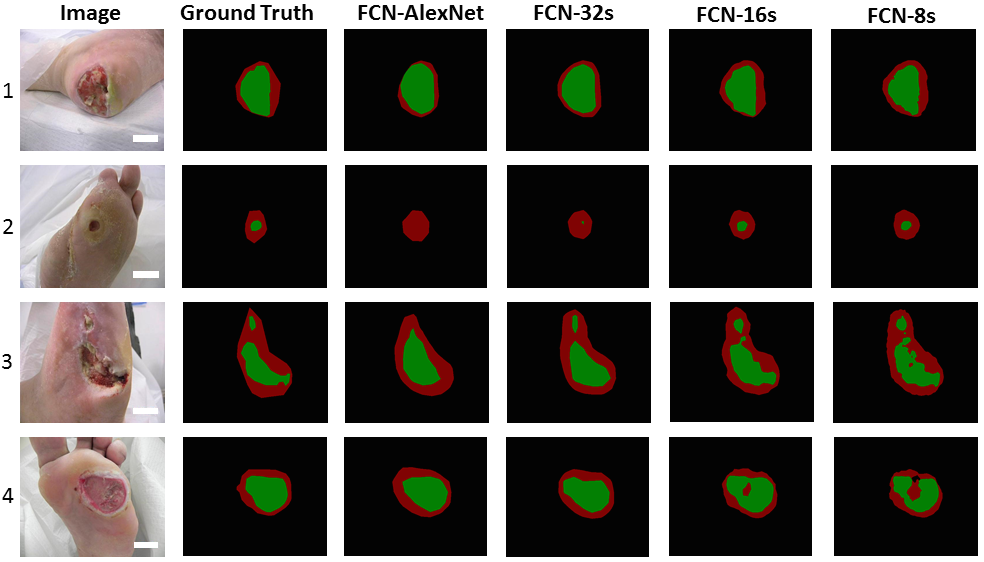}
	\caption{Four Examples of DFU and surrounding skin segmentation with the help of four different FCN models where green colour denotes the DFU area and the red denotes the surrounding skin.}
	\label{fig:Results}
\end{figure*}

In Table \ref{tab:tradFeats}, we report \textit{Dice Similarity Coefficient (Dice)}, \textit{Sensitivity}, \textit{Specificity}, \textit{Matthews Correlation Coefficient (MCC)} as our evaluation metrics for segmentation of DFU region.  In medical imaging, \textit{Sensitivity} and \textit{Specificity} are considered reliable evaluation metrics and where as for segmentation evaluation, \textit{Dice} are popularly used by researchers. 
\begin{equation}
TP= {|MP_1 \wedge GT_1|}
\end{equation}

\begin{equation}
FP= {|MP_1| - TP}
\end{equation}

\begin{equation}
FN= {|GT_1| - TP}
\end{equation}

\begin{equation}
TN= \forall - {|MP_1 \vee GT_1|}
\end{equation}

\begin{equation}
Sensitivity= \frac {TP}{TP+FN}
\end{equation}

\begin{equation}
Specificity= \frac {TN}{FP+TN}
\end{equation}

\begin{equation}
Dice= \frac {2*TP}{(2*TP + FP + FN)}
\end{equation}

\begin{small}
	\begin{equation}
	MCC= \frac {TP*TN - FP*FN}{\sqrt{(TP+FP)(TP+FN)(TN+FP)(TN+FN)}}
	\end{equation}
\end{small}

\noindent where MP is model predictions by various FCNs and GT is the ground truth labels.

\begin{table*}[]
	\centering
	\small\addtolength{\tabcolsep}{2pt}
	\renewcommand{\arraystretch}{1.5}
	\caption{Comparison of different FCNs architectures on DFU dataset (SS denotes Surrounding Skin)}
	\label{tab:tradFeats}
	\scalebox{0.85}{
		\begin{tabular}{|c|c|c|c|c|c|c|c|c|c|c|c|c|c}
			\cline{1-13}
			\multirow{2}{*}{Method} & \multicolumn{3}{c|}{\textit{Dice}}                              & \multicolumn{3}{c|}{\textit{Specificity}}                       & \multicolumn{3}{c|}{\textit{Sensitivity}}                       & \multicolumn{3}{c|}{\textit{MCC}}                               &  \\ \cline{2-13}
			& Complete         & Ulcer            & SS               & Complete         & Ulcer            & SS               & Complete         & Ulcer            & SS               & Complete         & Ulcer            & SS               &  \\ \cline{1-13}
			FCN-AlexNet             & 0.869            & 0.707            & 0.685            & 0.985 	& 0.982            & 0.991                     & 0.879    & 0.714            & 0.731                       & 0.859            & 0.697            & 0.694            &  \\ \cline{1-13}
			FCN-32s                 & \textbf{0.899}            & 0.763            & 0.762           & 0.989  & 0.986            & 0.989                     & \textbf{0.904}  & 0.751            & 0.823             & \textbf{0.891} & 0.752            & 0.768            &  \\ \cline{1-13}
			FCN-16s                 & 0.897 & \textbf{0.794} & \textbf{0.851} & 0.988 		& 0.986            & \textbf{0.994}             & 0.900 & \textbf{0.789} & \textbf{0.874}            & 0.889            & \textbf{0.785} & \textbf{0.852} &  \\ \cline{1-13}
			FCN-8s                  & 0.873            & 0.753            & 0.835           & \textbf{0.990}  & \textbf{0.987} & 0.993           & 0.854  	 & 0.726            & 0.847                      & 0.865            & 0.744            & 0.838            &  \\ \cline{1-13}
	\end{tabular}}
\end{table*}

In performance measures, FCN-16s was the best performer and FCN-AlexNet emerged as the worst performer for various evaluation metrics among all the other FCN architectures. Though, FCN architectures achieve comparable results when the evaluation is considered in the complete region. But, there is a notable difference in the performance of FCN models when ulcer and especially surrounding skin regions are considered. FCN-16s has achieved the best score of 0.794 ($\pm$0.104) in the ulcer region and 0.851 ($\pm$0.148) in the surrounding skin region for \textit{Dice}. whereas the FCN-32s achieved the best score 0.899 ($\pm$0.072) in the complete area determination. Overall, the FCN models has very high \textit{Specificity} for all the regions. Further, assessing the FCNs performance, we observed that FCN-16s and FCN-32s are better in \textit{Sensitivity}. FCN-16s performed best in the ulcer and surrounding skin regions and FCN-32s has the best in complete region performance in segmenting the complete region in terms of \textit{Sensitivity}, \textit{Dice} and \textit{MCC}. The results in Table \ref{tab:tradFeats} showed that the complete region segmentation has better performance than ulcer and surrounding skin in terms of \textit{Dice} and \textit{MCC}.

\begin{figure*}
	\centering
	\includegraphics[scale=0.5]{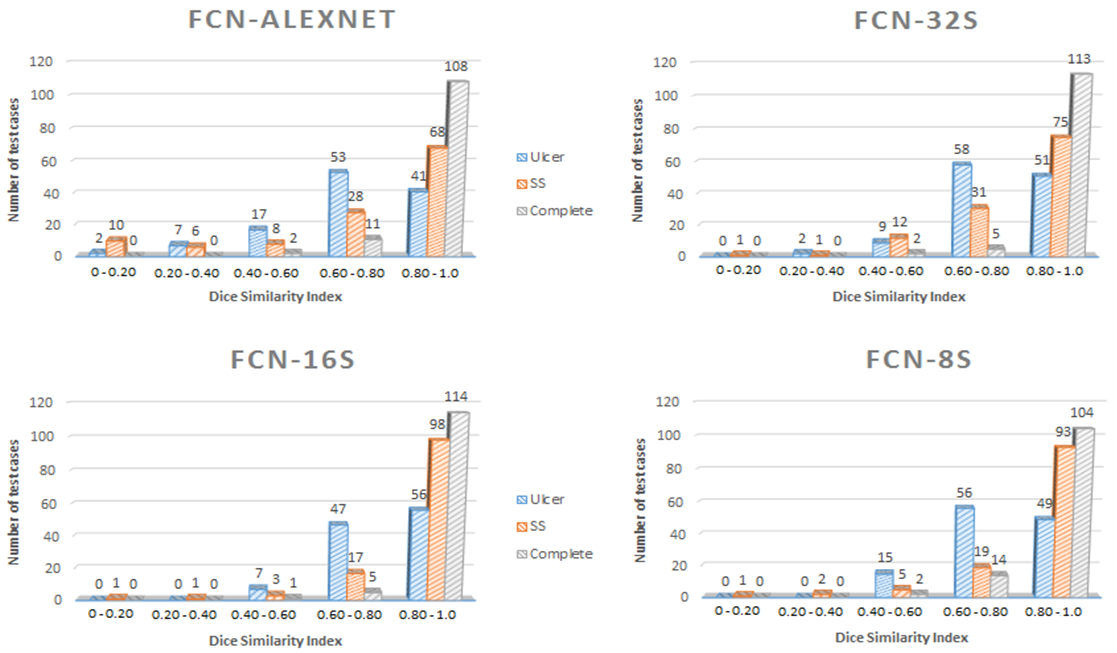}
	\caption{Distribution of Dice Similarity Coefficient for each trained model}
	\label{fig:Resultsgraph}
\end{figure*}
Finally, we tested the performance of the trained models on healthy foot images, they produced the highest specificity of 1.0 where neither DFU nor surrounding skin was detected.


\subsection{Inaccurate segmentation cases in FCN-AlexNet, FCN-32s, FCN-16s, FCN-8s}
Although the results are promising, there are few inaccurate segmentation cases that achieve very low \textit{Dice} score for each trained model as shown in Fig. \ref{fig:Resultsgraph}. In the Fig. \ref{fig:Results}, there are few instances in which FCN-AlexNet and FCN-32s models are not able to detect the small DFU and distinct surrounding skin or detect very small part of them. As discussed earlier, DFU and surrounding skin regions have very irregular outer boundaries, FCN-AlexNet and FCN-32s always tend to draw more regular contour and struggled to draw irregular boundaries to perform accurate segmentation, whereas, FCN-16s and FCN-8s with smaller pixel stride were able to produce more irregular contours of both DFU and surrounding skin. But, in few test images, some part of both categories overlap in some region due to the distinct tissues of DFU looks like surrounding skin and vice versa.

\section{Conclusion}
In this work, we developed deep learning approaches to train various FCNs that can automatically detect and segment the DFU and surrounding skin area with a high degree of accuracy. These frameworks will be useful for segmenting the other skin lesions such as moles and freckles, spotting marks (extending the work by Alarifi et. al. \cite{alarifi2017facial}), pimples, other wound pathologies classification, infections like chicken pox or shingles. This work also lays the foundations for technology that may transform the detection and treatment of DFU. This work has been done to achieve future targets that include: 1) to determine the various pathologies of DFU as multi-class classification and segmentation; 2) developing the automatic annotator that can automatically delineate and classify the DFU and related pathology; 3) developing various user-friendly system tools including mobile applications for DFU recognition and segmentation \cite{yap2015computer, yap2017footsnap} and computer vision assisted remote telemedicine system for the detection of DFU and provide feedback for different pathologies of diabetic feet. Moreover, this research could be applied to other related medical fields, for example, breast ultrasound lesions segmentation \cite{yap2007fully}.




%
\bibliographystyle{IEEEtran}
\bibliography{smc17}

\end{document}